\pdfoutput=1
\documentclass[runningheads]{llncs}

\usepackage{eccv}

\usepackage{eccvabbrv}
\usepackage{multirow}
\usepackage{graphicx}
\usepackage{subcaption}
\graphicspath{{figures/}{../figures/}}
\usepackage{booktabs}
\usepackage{amsmath}
\usepackage{amssymb}
\usepackage{array}
\usepackage{pdflscape}
\usepackage[accsupp]{axessibility}

\usepackage[pagebackref,breaklinks,colorlinks,citecolor=eccvblue]{hyperref}
\usepackage[nameinlink,capitalize]{cleveref}
\AtBeginDocument{\crefname{table}{Table}{Tables}\Crefname{table}{Table}{Tables}}
\usepackage{orcidlink}

\newcommand{\suppAppendix}{\cref{sec:appendix}}
\newcommand{\suppPayload}{\cref{sec:app-payload}}
\newcommand{\suppData}{\cref{sec:app-data}}

\newcommand{\suppIncrement}{\cref{sec:app-increment}}
\newcommand{\suppStaticMask}{\cref{sec:app-staticmask}}
\newcommand{\suppEthics}{\cref{sec:app-ethics}}

\begin{document}

\title{A Privacy Study of Sparse Collaborative Inference}

\titlerunning{A Privacy Study of Sparse Collaborative Inference}

\author{
Maximilian Andreas Hoefler$^{1}$ \thanks{Correspondence to maximilian.andreas.hoefler@hhi.fraunhofer.de}  \quad
Karsten Mueller$^{1}$ \quad
Wojciech Samek$^{1,2,3}$ 
}
\authorrunning{Hoefler et al.}
\institute{Fraunhofer Heinrich Hertz Institute, Germany \and Technical University of Berlin, Germany \and Berlin Institute for the Foundations of Learning and Data (BIFOLD) }

\maketitle

\begin{abstract}
Collaborative inference (CI) splits a model between an edge device and a
server, whereby the client computes an intermediate activation, transmits it, and
the server completes the computation. This raises two concerns, the
communication cost of the transmission and the risk that it reveals private
information about the input. Recent work reduces this cost by sparsifying
activations and entropy-coding the result. Sparsity has also been argued to
improve privacy, on the intuition that transmitting fewer values reveals
less about the input. We test this claim by decomposing the sparse
activation into the retained values and the set of positions they occupy,
and by reconstructing inputs from each component in isolation. We find that
sparsification reduces the leakage far less than it reduces the transmission
cost, and that the remaining risk shifts to the positions, which prior
analyses treat as side information for decoding. Across natural-image and face datasets, the
positions alone constitute a serious privacy risk, enabling high-fidelity
reconstructions and re-identification of individuals. The leakage from the
positions persists even when both the transmission cost and the task
utility are low. We conclude that the positions of sparse activations
should be treated as sensitive transmitted data and audited carefully in
the context of collaborative inference. Code is available at \url{https://github.com/an7123/Privacy-Study-Sparse-CI}. 

\end{abstract}

\section{Introduction}
\label{sec:intro}

Collaborative inference (CI) splits a model between a compute-constrained edge device
and a server. The client computes an intermediate activation and sends it to the
server, which completes inference~\cite{kang2017neurosurgeon,teerapittayanon2017distributed}.
This framework reduces the computational burden on the client, but it can introduce
a substantial communication cost, as intermediate activations are often larger
than the input itself. Much of the CI literature is therefore concerned with reducing
the transmitted payload through feature compression, learned bottlenecks,
quantization, or
sparsification~\cite{google_end_to_end,deepfeaturecompress,qunat_intermediate}.

A second concern in collaborative inference is privacy. Because the raw input never
leaves the device, CI is often described as privacy
preserving~\cite{kang2017neurosurgeon}, yet intermediate activations are not private
by construction, and prior work has repeatedly shown that inputs can be recovered
from transmitted features~\cite{dosovitskiy2016inverting}. A considerable literature on CI defenses accordingly
seeks to reduce this leakage by perturbing, pruning, or adversarially training the
transmitted
representation~\cite{hoefler2026leveraging,titcombe2021practical,ding2024patrol,nopeek,CEM}.
Within this literature, sparsity has been suggested as a mechanism that addresses
communication and privacy simultaneously, since transmitting fewer entries reduces
bandwidth and exposes a smaller attack
surface~\cite{hoefler2026leveraging,hoefler2025relevance,scheliga2022precode,gradient_defense_sparse}.
The suggestion is attractive because sparsity additionally aligns with standard
codecs, which exploit zeros and structured masks for efficient
compression~\cite{becking2023nncodec}.

Beyond qualitative arguments, however, the privacy consequences of
sparsification itself have received little direct study. In this work we show that
sparsification does more than reduce the size of the transmitted message, it
restructures the object that an adversary observes and thus changes the
privacy risk. A sparse representation consists of two components, both of which an adversary
can exploit: the retained \emph{values} and the \emph{positions} at which those
values occur. Existing privacy
analyses of collaborative inference concentrate on the values, which are the
natural target of compression~\cite{CEM,ding2024patrol},
perturbation, and quantization, whereas the positions are usually treated as side
information for decoding. The positions, however, are also input-dependent and can therefore
reveal information about the client's data even after every value has been erased. The
objective of this work is to determine not only how much information survives
sparsification, but which component of the sparse activation carries the information that
privacy attacks require.

We structure our work around four central questions. The first (RQ1) is whether
the leakage resides in the values or the positions. To separate them we
reconstruct the input from each component in isolation (\cref{fig:split}): one
probe sends only the binary mask of retained positions, the other sends the
retained values on random positions. Because the two exchange exactly one
component, their comparison identifies the one responsible for the leakage.

A second question (RQ2) is whether the privacy risk decreases as fast as the
transmission cost when the activation is made sparser. Answering this question
requires a common measure of that cost, for which we use the rate, defined as
the number of bits needed to transmit the sparsified activation per activation dimension. This rate consists of two parts, the bits
that encode which positions are retained and the bits that encode the values
placed at those positions. We find that the positions remain a serious privacy
risk even at low bit rates.

A third question (RQ3) is whether sparsification reduces the re-identification
risk specifically, since identity is often assumed to reside in the fine detail
that sparsification removes. We score visual and biometric leakage separately
(SSIM, and FaceNet similarity with rank-1/rank-5 retrieval) and find the
positions alone re-identify individuals far above chance even when fine visual
detail is lost, which, because such reconstructions can fall within the scope
of regulation such as the GDPR, is a compliance concern as well as a technical
one.

Finally, we study the strength of the threat model and ask which attacker poses the strongest risk (RQ4). We invert every probe under two adversaries, namely
a white-box audit that optimizes for an input whose activation matches the observation
using the model alone, and an auxiliary-data attacker that trains a
decoder on images from a related distribution, modeling a server that holds data
similar to the client's. Across datasets, architectures, and settings, the
auxiliary-data attacker proves the more consequential of the two, whereas the
white-box audit is effective mainly against representations that transmit
continuous values.

We offer the following contributions:
\begin{itemize}
\item We introduce a probing framework that attributes reconstruction leakage to a
single component of a sparse activation. The framework decomposes the activation into
values and positions, transmits each component in isolation, and
adds corruption controls that measure how much of the positional structure an
adversary requires.
\item We provide a systematic measurement of sparse-representation privacy leakage across sparsity
levels, datasets, split layers, and backbones, under both an optimization audit and
a learned inverse, with visual and biometric leakage scored separately; the
positions account for most of the measured leakage, including identity, while
the values add only a small increment on the correct support and nothing
without it.
\item We show that the privacy risk persists at operating points where both the
transmitted rate and the task utility are low. Rate and leakage are decoupled, in
that sparsification reduces the rate far faster than the leakage at matched
sparsity and at matched utility alike.
\item We derive consequences for privacy auditing. The standard white-box audit
underreports the leakage from the positions relative to an attacker with auxiliary data. Audits of
sparse collaborative inference should therefore include learned inversion attacks,
and the input-dependent positions should be treated as sensitive transmitted data
rather than as harmless side information.
\end{itemize}

\section{Related Work}
\label{sec:related}

\subsubsection{Compression for collaborative inference.}
Because split-layer tensors can exceed the input in payload, CI systems compress
the transmitted features, e.g.\ via intermediate reduction~\cite{shao2020bottlenet},
learned end-to-end features~\cite{hao2022multiagent}, or sparsification such as
relevance-guided static masking with entropy
coding~\cite{hoefler2025relevance,becking2023nncodec}. 

\subsubsection{Feature inversion.}
Feature inversion measures what a representation reveals, by optimizing a
matching image~\cite{mahendran2015understanding} or learning an inverse from
features to images~\cite{dosovitskiy2016inverting}; in CI, model-inversion
attacks train such an inverse on auxiliary
data~\cite{he2019modelinversion,he2020attacking}. Our two attackers instantiate
these families component-wise on the sparse activation. That a support alone can
suffice for recovery has precedent in one-bit compressed
sensing~\cite{boufounos2008onebit}; we measure how much trained CI
representations leak through this channel in practice.

\subsubsection{Privacy defenses for CI.}
Defenses against reconstruction include noise injection into intermediate
features~\cite{titcombe2021practical}, attacker-aware training that penalizes a
simulated adversary~\cite{li2022ressfl,ding2024patrol}, and sparsity-based
obfuscation~\cite{zheng2023randomized,hoefler2026leveraging}. Some methods also rely on
information-theoretic tools such as~\cite{CEM}, which give a lower bound on the privacy
leakage. Closest to our study, \cite{hoefler2026leveraging} formulates sparse
activations as separate index and value channels and proposes protecting each
with an information-theoretic budget. Our goal is complementary: we directly
measure what an adversary can recover from each channel in isolation. By
transmitting only the input-dependent support, or retaining the values while
randomizing their positions, we show that the support accounts for most
reconstruction and re-identification leakage. We test this finding across
attackers, sparsity levels, datasets, split layers, and backbones.

\section{Study Design}
\label{sec:design}
The design follows the research questions of \cref{sec:intro}. We split the transmitted
sparse activation into values and positions, and construct probes that isolate which component is responsible for the leakage and how the leakage changes as the activation is made sparser (RQ1 and RQ2, \cref{sec:decomp}). We then invert every probe under two attackers, a white-box audit available to a defender and a server that holds auxiliary data (RQ4, \cref{sec:threat}).

\subsection{Setup and Notation}
\label{sec:setup-notation}
We study a split-inference system in which a network $f = f_{\mathrm{s}} \circ
f_{\mathrm{c}}$ is partitioned at a chosen layer, where $f_{\mathrm{s}}$ is the server
model and $f_{\mathrm{c}}$ the client model. On input $x$ the client evaluates
\begin{equation}
z = f_{\mathrm{c}}(x) \in \mathbb{R}^{d},
\end{equation}
transmits $z$, and the server completes the prediction
$\hat{y} = f_{\mathrm{s}}(z)$. All splits in this work are taken after a ReLU,
so $z \ge 0$; the retained values are therefore non-negative, and values and
magnitudes coincide. Our goal is to determine how an attacker can invert each component to reconstruct $x$.

\subsection{Separating Values and Positions of Sparse Activations}
\label{sec:decomp}
\begin{figure}[t]
\centering
\includegraphics[width=0.8\linewidth]{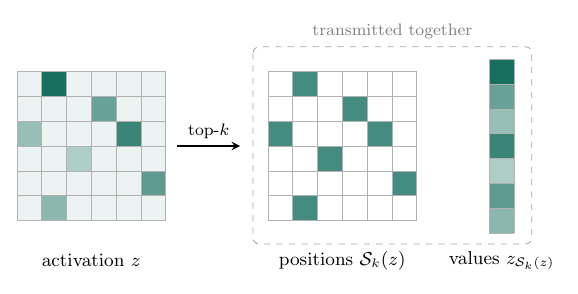}
\caption{Decomposition of a sparse activation. Top-$k$ selection splits the
activation $z$ into the transmitted positions, a binary mask over the $k$
largest-magnitude coordinates, and the transmitted values at those
coordinates. The two components are normally transmitted together, but they
can be inverted separately, which is the basis of our probing framework.}
\label{fig:split}
\end{figure}

Let $\mathcal{S}_k(z) \subseteq \{1,\dots,d\}$ index the $k$ entries of $z$ with
largest magnitude, equivalently the super-level set $\{i : |z_i| \ge \tau_k\}$ for
the rank-$k$ threshold $\tau_k$. We write $\rho = k/d$ for the fraction of retained
units, so the sparsity level is $1-\rho$. A top-$k$ sparse activation is the pair
\begin{equation}
\mathcal{T}_k(z) = \bigl(\,\underbrace{\mathcal{S}_k(z)}_{\text{positions}},\ \,
\underbrace{z_{\mathcal{S}_k(z)}}_{\text{values}}\,\bigr),
\label{eq:decomp}
\end{equation}
the discrete set of \emph{positions} $\mathcal{S}_k(z)$, in the literature also
called the support, together with the continuous \emph{values} at those
coordinates (\cref{fig:split}). We treat the two parts as separately transmittable objects and use them as the basis of a probing
framework that measures the privacy leakage of each.

\begin{table}[t]
\centering
\caption{Transmitted representations as a two-factor design (source of the
positions $\times$ whether values are sent, at precision $b$); rates in
bits/dim from \cref{eq:rate}. ``--'' cells are unused (see \cref{sec:decomp}).}
\label{tab:repr-map}
\footnotesize
\setlength{\tabcolsep}{4pt}
\renewcommand{\arraystretch}{1.0}
\begin{tabular}{@{}>{\raggedright\arraybackslash}p{0.20\linewidth}
                   >{\raggedright\arraybackslash}p{0.30\linewidth}
                   >{\raggedright\arraybackslash}p{0.42\linewidth}@{}}
\toprule
positions\,$\downarrow$\ \big/\ values\,$\rightarrow$
 & \textbf{No values} {\scriptsize(mask, $b{=}0$)}
 & \textbf{Real values} {\scriptsize(precision $b$)} \\
\midrule
\textbf{All positions} {\scriptsize(no sparsity)}
 & --
 & Dense {\scriptsize($b{=}16$, rate $16$)}; $b$-bit quant.\ {\scriptsize($b{\in}\{1,2,4,8\}$, rate $b$)} \\
\textbf{True top-$k$} {\scriptsize(input-dep.)}
 & Positions only {\scriptsize(rate $H_2(\rho)$)}
 & Top-$k$ activation {\scriptsize(FP16, rate $H_2(\rho){+}16\rho$)} \\
\textbf{Block-shuffled} {\scriptsize($4{\times}4$)}
 & Block-shuffled positions
 & -- \\
\textbf{Random} {\scriptsize(input-indep.)}
 & Random positions
 & Values on random positions \\
\bottomrule
\end{tabular}
\end{table}

\subsubsection{Rate accounting.}
Having split the sparse activation into values and positions, we assign each
component a bit cost, so that in \cref{sec:results} the leakage of a
component (SSIM, re-identification) can be compared against the rate it
consumes. Because the transmitted activation is a deterministic function of the input, a
component's bit cost also bounds the information it can carry about it.

The two components cost different amounts. The $\binom{d}{k}$ position sets need
at most $\log_2\binom{d}{k}$ bits (correlated masks compress further,
\cref{sec:limitations}), while the values add $b$ bits per retained coordinate,
where $b$ is the value precision. Normalizing by $d$ with
$\log_2\binom{d}{k}\approx d\,H_2(\rho)$ and $H_2$ the binary entropy, the rate
per activation dimension is
\begin{equation}
R \;=\; \underbrace{H_2(\rho)}_{\text{positions}}
\;+\; \underbrace{\rho\, b}_{\text{values}}
\qquad \text{bits/dim}.
\label{eq:rate}
\end{equation}
\Cref{tab:repr-map} lists the resulting rate for each different representation (detailed in the next section).
Positions-only sets $b=0$ meaning the mask is transmitted, but no bits describe
the values placed on it. Top-$k$ uses an FP16 encoding of the values ($b=16$), and the
dense references have $\rho=1$. 

\subsubsection{Probes.}
Having split the activation into values and positions, we now
define the probes that transmit them in isolation. \cref{tab:repr-map} organizes
them by two factors: the source of the positions, and whether the values are
sent.

Two probes carry the central contrast for RQ1. \emph{Positions-only}
transmits the binary mask $m \in \{0,1\}^d$ with
$m_i = \mathbb{1}[\,i \in \mathcal{S}_k(z)\,]$, i.e., the input-dependent
positions, no values. \emph{Values-on-random-positions} is its complement:
the retained values are placed on a freshly drawn index set of the
same size, redrawn per sample, so the values are preserved while the
positions carry no information about the input. Comparing the two
identifies the privacy risk of each component. If the values were responsible,
relocating them to random positions would preserve the leakage and
positions-only would collapse; if the positions were responsible, the
reverse would hold.

The reference points (dense activation, top-$k$ activation, and dense $b$-bit
quantization, $b\in\{1,2,4,8\}$) bound the leakage of the full layer and the full
sparse activation. The corruption controls measure how much positional structure
the leakage needs: the block-shuffled control applies a per-sample $4\times4$
block permutation to the mask, and the random control
is a matched-density input-independent mask.

\subsection{Threat Models}
\label{sec:threat}
We consider two attackers. The first has access only to the model. The second additionally
holds auxiliary data drawn from the same or a similar distribution as the client data. Both
observe $o = \mathcal{O}(f_{\mathrm{c}}(x))$, where $\mathcal{O}$ is the transmitted-probe
operator from \cref{tab:repr-map}, for example top-$k$ selection, the position mask, or the
identity map for dense activations. The two attackers differ only in the availability of
auxiliary data.

\subsubsection{Adaptive white-box inversion.}
Without auxiliary data, the attacker reconstructs the input by optimization,
\begin{equation}
\hat{x} \;=\; \arg\min_{x'}\; \ell\bigl(\mathcal{O}(f_{\mathrm{c}}(x')),\, o\bigr)
\;+\; \lambda\,\Omega(x'),
\label{eq:whitebox}
\end{equation}
where $\Omega$ is an image prior and $\ell$ is a fidelity term matched to the observation.
For continuous channels, $\ell$ matches the retained values at the transmitted positions,
$\lVert f_{\mathrm{c}}(x')_{\mathcal{S}} - z_{\mathcal{S}}\rVert^2$. For a positions-only mask,
$\ell$ matches the induced positions.

\subsubsection{Auxiliary-data inversion.}
The second attacker holds samples $\{x_j\}$ from the same or a related distribution,
possibly the training data of the model, and learns the inverse mapping. From pairs
$\bigl(\mathcal{O}(f_{\mathrm{c}}(x_j)),\,x_j\bigr)$ it trains a convolutional decoder
$g_\phi$ by minimizing
\begin{equation}
\min_{\phi}\; \sum_{j}\bigl\lVert g_\phi\bigl(\mathcal{O}(f_{\mathrm{c}}(x_j))\bigr)
- x_j\bigr\rVert^2,
\label{eq:auxdec}
\end{equation}
and reconstructs a held-out observation as $\hat{x}=g_\phi(o)$ with
$o=\mathcal{O}(f_{\mathrm{c}}(x))$. This attacker represents the realistic server threat
when related data is available. It is the relevant model for discrete positions, because a
learned inverse can capture the distributional regularities that map a binary mask to an
image, whereas a hand-designed objective cannot represent them. The auxiliary
set need not be large or class-matched: with as few as $100$ images the learned
inverse already surpasses the white-box audit, and it transfers across disjoint
classes.

\subsubsection{A values-matched attacker.}
Because a convolutional inverse suits a mask but not an unordered set of values,
a collapse on values-on-random-positions might reflect an architecture mismatch
rather than an absence of information. As a control (\cref{sec:res-topology}) we
additionally invert the raw values with two permutation-invariant attackers (a
sorted-value MLP and DeepSets), so that
a failure to reconstruct cannot be blamed on capacity; construction details are
in \suppPayload.

\section{Results}
\label{sec:results}
\subsection{Experimental Setup}
\label{sec:setup}
Our default network is ResNet-18~\cite{he2016resnet} which we split at \texttt{layer2} and evaluate on three datasets.
TinyImageNet provides natural images, Imagenette is a $10$-class
ImageNet subset at higher resolution, included to test whether the results depend on the
pixel resolution. FaceScrub~\cite{ng2014facescrub} is a face dataset,
for which reconstruction corresponds to an identity risk.

We report task utility, rate, and two families of leakage metrics. We separate visual from
biometric leakage because pixel fidelity and identity are not equivalent and need not vary
together (RQ3). Utility is the downstream top-1 accuracy of the server on the transmitted
representation. Visual leakage is the SSIM between $x$ and $\hat{x}$. For FaceScrub,
biometric leakage is the FaceNet~\cite{schroff2015facenet} cosine similarity
$\cos(\varphi(\hat{x}),\varphi(x))$ between embeddings, together with rank-1 and rank-5
re-identification against an external gallery that holds one held-out image per identity, so
that retrieval cannot succeed by matching the query image itself. Rate is the analytical
estimate from \cref{eq:rate}, in bits per activation dimension.

\subsubsection{Implementation details.}
The Aux attacker is a convolutional decoder whose input head adapts to the transmitted
object and whose generative trunk is identical across probes ($1.78$\,M parameters),
trained with Adam for $12$ epochs on TinyImageNet and $20$ on FaceScrub. The WB audit
minimizes \cref{eq:whitebox} with a total-variation prior, using Adam on the image for
$1500$ to $2000$ steps. Architecture, capacity, and optimization budget are identical
across all probes, so no representation is favored by attacker capacity. Full
configurations, dataset partitions, and parameter counts are given in \suppAppendix.

Throughout this work, WB denotes the white-box audit attacker and Aux the
auxiliary-data attacker of \cref{sec:threat}; probe names follow
\cref{tab:repr-map}. Unless stated otherwise, results are at the main
operating point of $95\%$ sparsity. Unless stated otherwise, reported values
are the mean over five independent runs, and $\pm$ denotes the standard
deviation across those runs.

\subsection{The Positions Carry the Leakage}
\label{sec:res-topology}
We first study whether the privacy leakage stems from positions or values to determine which
component of the sparse activation carries the leakage (RQ1). \Cref{tab:combined95}
reports the comparison under both white-box (WB) and auxiliary (Aux) attackers. On TinyImageNet under the Aux
attacker, positions-only transmission recovers almost as
much of the input as the full top-$k$ activation ($0.428$ versus $0.435$ SSIM). The complementary probe,
values on random positions, falls to the random-positions floor in both
reconstruction ($0.145$ versus $0.142$ SSIM) and utility ($16.2\%$ versus
$15.5\%$). This pattern identifies the positions as the
component that accounts for most of the measured leakage, since removing the
values barely changes it, whereas deleting the input-dependence of the
positions removes it. The values are not strictly uninformative given the
positions, but their contribution is small and conditional on the correct
support, and vanishes without it (\suppIncrement). We find the positions are
sufficient for most of the leakage, not the sole possible source of it. The
collapse of the
values probe is moreover not an artifact of the attacker architecture, since the
values-matched attackers of \cref{sec:threat} likewise recover no more than the
unconditional mean image ($0.144$ versus $0.145$ SSIM). Sparsification does
reduce leakage relative to the dense activation ($0.559$ to $0.428$ SSIM under
Aux), but by far less than it reduces the rate (\cref{sec:res-rate} quantifies
this gap).



The comparison transfers across datasets and resolution. On Imagenette, a
higher-resolution setting, positions-only again nearly matches the full top-$k$ activation
($0.400$ versus $0.420$ SSIM, \cref{tab:combined95}), and replacing the
input-dependent positions with random ones drops downstream accuracy from
$82.3\%$ to near chance, so the positions are task-relevant beyond $64$-pixel
inputs. On faces, the same positions-only reconstructions carry identity, which
we examine in \cref{sec:res-biometric}. Qualitative reconstructions
(\cref{fig:recon}) show the same ordering: positions-only tracks the full
top-$k$ activation across datasets, while values on random positions reduce to a
dataset prior.

\begin{table*}[t]
\centering
\caption{Reconstruction leakage at 95\% sparsity on TinyImageNet and
Imagenette. For each dataset we report downstream accuracy (Acc.,
attack-independent) and reconstruction SSIM under the WB audit and the Aux
inverse (\cref{sec:threat}).}
\label{tab:combined95}
\setlength{\tabcolsep}{5pt}
\resizebox{\textwidth}{!}{%
\begin{tabular}{l ccc ccc}
\toprule
\multirow{2}{*}{Representation}
  & \multicolumn{3}{c}{TinyImageNet}
  & \multicolumn{3}{c}{Imagenette} \\
\cmidrule(lr){2-4}\cmidrule(lr){5-7}
                        & Acc.          & SSIM (WB)         & SSIM (Aux)        & Acc.          & SSIM (WB)         & SSIM (Aux)         \\
\midrule
  Dense activations           & $53.1{\pm}0.8$ & $0.436{\pm}0.001$ & $0.559{\pm}0.002$ & $88.9{\pm}0.4$ & $0.496{\pm}0.003$ & $0.561{\pm}0.003$ \\
  Top-$k$ activation          & $52.2{\pm}0.6$ & $0.213{\pm}0.008$ & $0.435{\pm}0.005$ & $86.0{\pm}0.6$ & $0.127{\pm}0.008$ & $0.420{\pm}0.005$ \\
  Positions only              & $49.8{\pm}0.5$ & $0.059{\pm}0.006$ & $0.428{\pm}0.006$ & $82.3{\pm}0.3$ & $0.013{\pm}0.009$ & $0.400{\pm}0.005$ \\
\midrule
  Values on rand.\ positions  & $16.2{\pm}0.4$ & $0.038{\pm}0.003$ & $0.145{\pm}0.003$ & $13.0{\pm}0.4$ & $0.041{\pm}0.004$ & $0.216{\pm}0.007$ \\
  Block-shuffled positions    & $18.6{\pm}0.9$ & $0.061{\pm}0.003$ & $0.295{\pm}0.004$ & $64.7{\pm}0.4$ & $0.052{\pm}0.003$ & $0.300{\pm}0.004$ \\
  Random positions            & $15.5{\pm}0.3$ & $0.036{\pm}0.003$ & $0.142{\pm}0.001$ & $12.4{\pm}0.4$ & $0.038{\pm}0.003$ & $0.214{\pm}0.002$ \\
\bottomrule
\end{tabular}}
\end{table*}

\subsection{Rate and Privacy Risk}
\label{sec:res-rate}
\begin{figure}[t]
\centering
\includegraphics[width=0.82\linewidth]{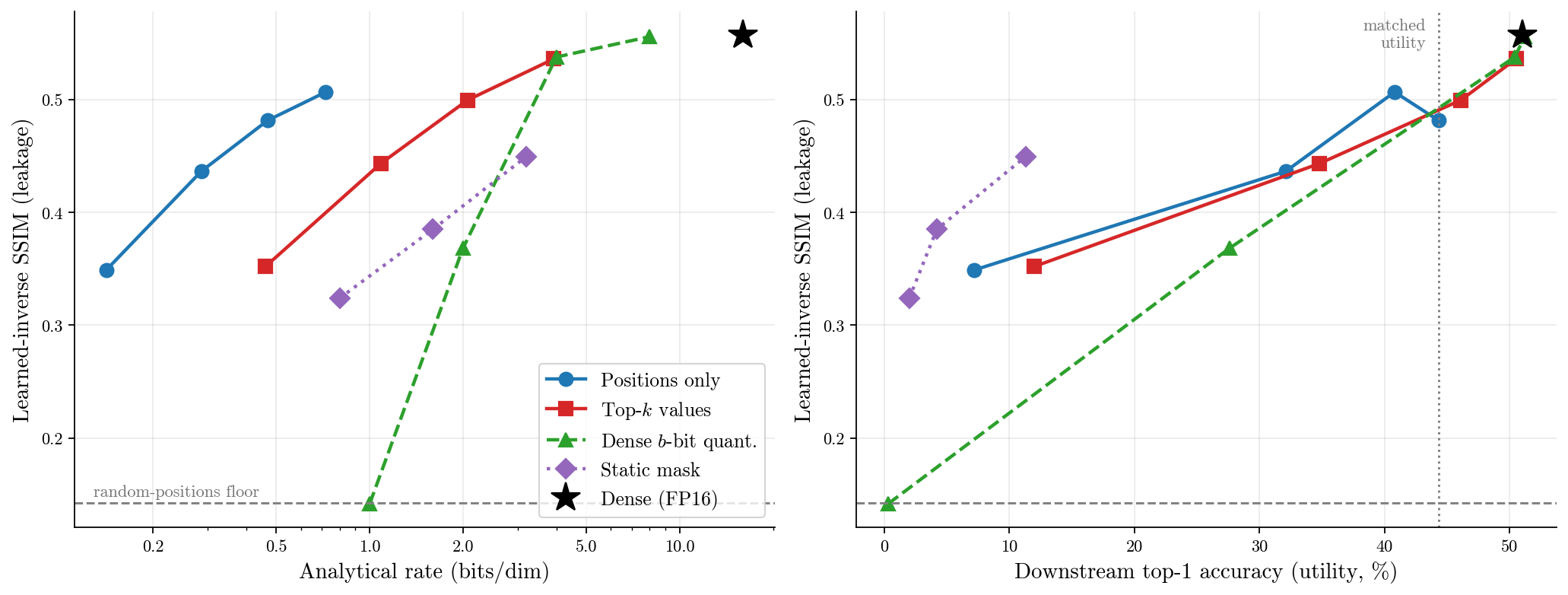}
\caption{Trade-off between rate, utility, and leakage on TinyImageNet.}
\label{fig:pareto}
\end{figure}

We next ask whether privacy risk, measured in terms of reconstruction,
decreases as fast as the rate (RQ2). \Cref{fig:pareto} sweeps positions-only and
the top-$k$ activation over keep-fractions, together with dense
$\{1,2,4,8\}$-bit quantization and a static relevance
mask~\cite{hoefler2025relevance}, and plots the accuracy of the frozen server
against reconstruction SSIM under the Aux inverse, at the analytical rate of
\cref{eq:rate}. Three comparisons structure the figure. Increasing sparsity reduces the rate and eventually the
utility, but the leakage decreases only gradually. Positions-only at $98\%$
sparsity costs $0.141$ bits per dimension yet still yields $0.349$ SSIM. The
reconstruction is not photorealistic (\cref{fig:recon}), but the
privacy-sensitive structure does not decrease in proportion to the bandwidth. At matched utility (approximately $44$ to $46\%$ accuracy), positions-only
leaks almost as much as the full top-$k$ activation ($0.481$ versus $0.499$ SSIM)
and only slightly less than dense $4$-bit quantization ($0.538$), at reduced rate. Sparsity therefore reduces the
rate far more than it reduces the leakage, at matched utility as well as at
matched sparsity. This implies that an activation can retain sensitive
information even when both its task utility and its transmission cost are
low. Finally, the input-independent static mask still transmits real values on its
fixed support and therefore still leaks (\cref{fig:pareto}). Input-independence
removes the \emph{position} channel, not the \emph{value} channel, and the mask
never reaches usable frozen-server accuracy
(\suppStaticMask). Wherever dynamic top-$k$ is chosen for its
compression and utility, its input-dependent positions belong within the
privacy threat model.

\subsection{The Positions Retain Biometric Information}
\label{sec:res-biometric}
RQ3 asks whether the positions leak identity in addition to pixel appearance,
since the two need not vary together. We reconstruct FaceScrub images from each
probe under the Aux inverse, embed the reconstruction with FaceNet, and retrieve
against a gallery that holds one held-out image for each of the $211$ evaluation
identities. Each query is the reconstruction of a different image of the same
identity, so a query image never appears in the gallery and retrieval cannot
succeed by matching a query to itself. Chance rank-1 is therefore
$1/211 = 0.47\%$. In this main protocol the inverse may have been trained on
other images of the evaluation identities, so identities can overlap between
decoder training and the gallery even though images never do. We report the
embedding cosine and rank-1/rank-5 retrieval (\cref{tab:facescrub}) and remove
the identity overlap in the control below. The full identity partition is given
in \suppData. One might expect identity to be lost with sparsification, as positions-only
reconstructions on TinyImageNet lose the high-frequency texture of the dense
activation (\cref{sec:res-rate}), and identity is also carried by fine facial
detail. However, we find that re-identification is still possible. For example,
positions-only reconstructions retrieve the correct identity at $24.5\%$ rank-1
and $49.5\%$ rank-5, close to the $24.9\%$ and $51.3\%$ of the full top-$k$
activation, whereas the values on random positions retrieve near the
random-positions floor at rank-1 ($1.1\%$ versus $0.6\%$, chance $0.47\%$) and
retain only a small residual signal at rank-5 ($11.6\%$, against a chance level
of approximately $2.4\%$). One may argue that the decoder could retrieve a memorized prototype of a known
identity rather than reconstruct the subject. We therefore repeat the experiment
with an identity-disjoint partition. The inverse is trained only on the $60\%$ of
identities that never appear at evaluation, and both the queries and the gallery
are drawn from the disjoint $40\%$, the $211$ held-out identities, with a fresh
partition per seed and results averaged over five seeds. The decoder has
therefore never seen any evaluated identity, and chance rank-1 stays at about
$0.47\%$. This costs about $3.7$ rank-1 points. Positions-only still
re-identifies unseen subjects at $20.8\%$ rank-1, approximately $44$ times
chance, and $45.3\%$ rank-5, well above the random-positions floor.

\begin{table*}[t]
\centering
\caption{FaceScrub leakage at 95\% sparsity. Acc.\ is downstream accuracy,
SSIM is reconstruction fidelity under the WB audit and the Aux inverse
(\cref{sec:threat}); ID cos.\ is FaceNet embedding cosine and R@1/R@5 are
rank-1/rank-5 re-identification against an held-out evaluation gallery (one held-out
image per identity), both under the Aux inverse.}
\label{tab:facescrub}
\setlength{\tabcolsep}{5pt}
\resizebox{\textwidth}{!}{%
\begin{tabular}{l r cc r c}
\toprule
\multirow{2}{*}{Representation} & \multirow{2}{*}{Acc.}
  & \multicolumn{2}{c}{SSIM}
  & \multirow{2}{*}{ID cos.\ (Aux)} & \multirow{2}{*}{R@1/R@5 (Aux)} \\
\cmidrule(lr){3-4}
  &                & WB                & Aux                &                     &                                     \\
\midrule
  Dense activations           & $75.0{\pm}0.8$ & $0.516{\pm}0.003$ & $0.840{\pm}0.001$ & $0.623{\pm}0.004$  & $38.4{\pm}0.5$\,/\,$63.5{\pm}0.6$ \\
  Top-$k$ activation          & $74.0{\pm}0.6$ & $0.436{\pm}0.003$ & $0.754{\pm}0.008$ & $0.514{\pm}0.009$  & $24.9{\pm}0.6$\,/\,$51.3{\pm}0.6$ \\
  Positions only              & $63.0{\pm}0.6$ & $0.076{\pm}0.007$ & $0.751{\pm}0.002$ & $0.510{\pm}0.002$  & $24.5{\pm}0.9$\,/\,$49.5{\pm}0.7$ \\
\midrule
  Values on rand.\ positions  & $20.0{\pm}0.7$ & $0.154{\pm}0.004$ & $0.290{\pm}0.003$ & $0.039{\pm}0.008$  & $1.1{\pm}0.8$\,/\,$11.6{\pm}0.9$  \\
  Block-shuffled positions    & $50.8{\pm}0.6$ & $0.161{\pm}0.004$ & $0.501{\pm}0.002$ & $0.021{\pm}0.003$  & $1.1{\pm}0.8$\,/\,$12.0{\pm}0.4$  \\
  Random positions            & $18.0{\pm}0.8$ & $0.143{\pm}0.004$ & $0.276{\pm}0.001$ & $0.012{\pm}0.002$  & $0.6{\pm}0.1$\,/\,$2.6{\pm}0.8$   \\
\bottomrule
\end{tabular}}
\end{table*}

\begin{figure}[t]
\centering
\includegraphics[width=0.9\linewidth]{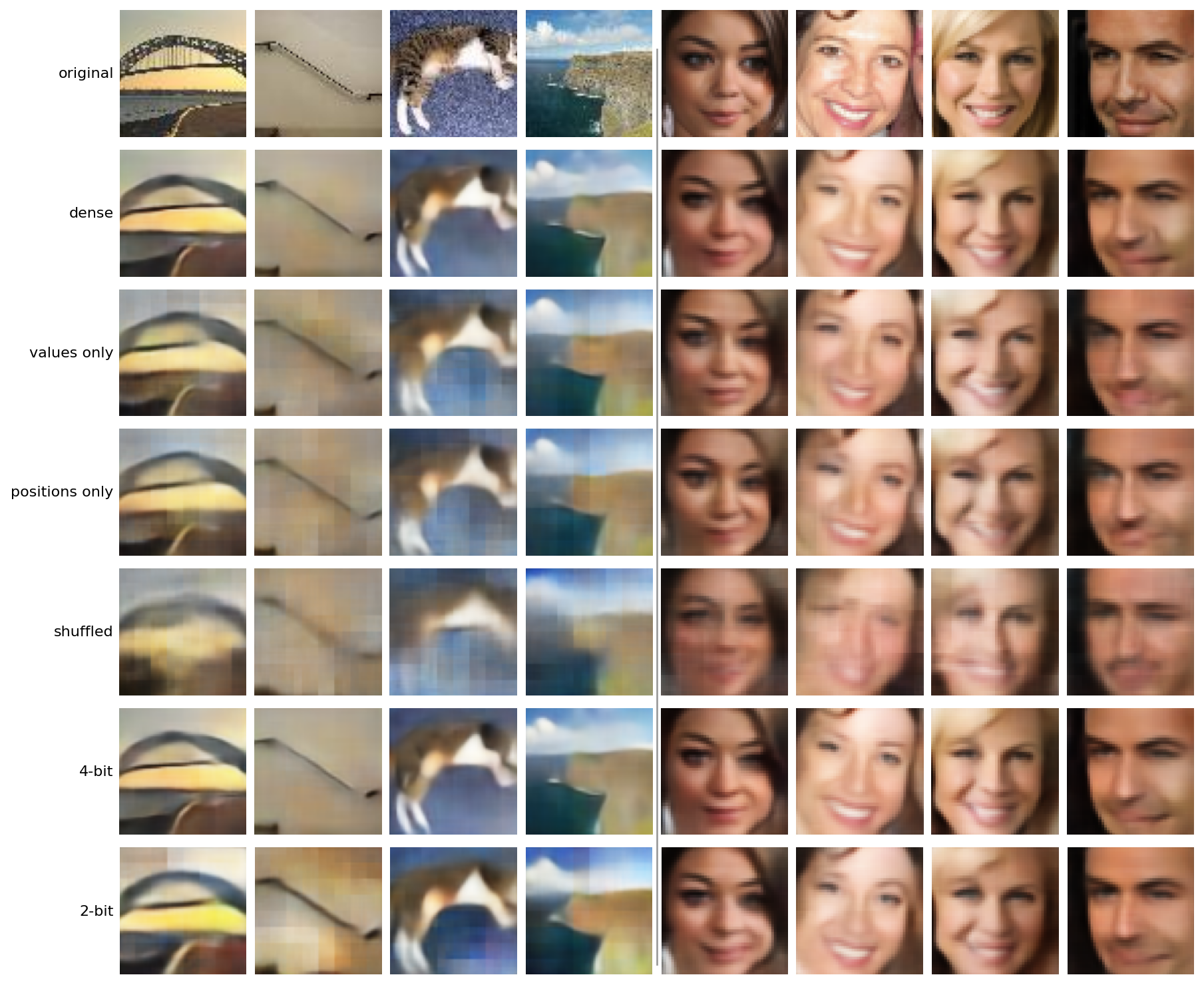}
\caption{Reconstructions under the Aux inverse for TinyImageNet (left four
columns) and FaceScrub (right four). Rows, top to bottom: original, dense
activation, the top-$k$ activation (the retained values at their true
positions, labeled ``values only'' in the figure), positions only,
block-shuffled positions, dense $4$-bit, dense $2$-bit.}
\label{fig:recon}
\end{figure}

\Cref{fig:recon} shows the effect on faces. Reconstructions from the correct
positions remain identity-specific and closely follow those from the full
top-$k$ activation, whereas block-shuffling the positions degrades them toward a
generic face. RQ4 asks whether the privacy conclusion depends on the attacker. We find that it does, and that the white-box audit, as commonly instantiated, is not the strongest inversion attack for a discrete representation. The WB and Aux columns of \cref{tab:combined95,tab:facescrub} compare the two attackers of \cref{sec:threat} on identical observations. For positions-only masks on FaceScrub, the WB audit reaches only $0.076$ SSIM, whereas the Aux decoder reaches $0.751$ SSIM and a FaceNet cosine of $0.510$; the WB audit has no identity entry in \cref{tab:facescrub}, as its positions-only reconstructions are too degraded to embed meaningfully. The gap is representation-specific rather than uniform: the same optimizer recovers the full top-$k$ activation well ($0.436$ SSIM on FaceScrub, $0.213$ on TinyImageNet) but not the positions ($0.076$ and $0.059$). The learned inverse exploits distributional regularities of the sparse mask that a hand-designed matching objective cannot represent, so the WB audit is poorly matched to the discrete mask observation. The corruption controls make the mismatch visible: under WB, the block-shuffled mask even scores above the true positions ($0.161$ versus $0.076$ SSIM on FaceScrub), inverting the Aux ordering, which further indicates that the audit's objective, not the representation's information content, limits WB performance. The Aux attack is also data-efficient and robust to distribution shift. At a fixed training budget, positions-only SSIM increases monotonically with the size of the auxiliary set, from $0.173$ at $100$ images, already well above the WB value of $0.059$, to $0.414$ at $100{,}000$ images. A decoder trained on a disjoint set of TinyImageNet classes transfers without loss ($0.405$ in-distribution versus $0.423$ on held-out classes). Stronger white-box variants, such as relaxed top-$k$ objectives, deeper image priors, or optimization within a trained generator's latent space, may narrow the gap and are left to future work.

\subsection{Robustness: Layers, Backbones, and Sparsity Structure}
\label{sec:res-ablations}
\begin{figure*}[t]
\centering
\begin{subfigure}{0.49\textwidth}
\centering
\includegraphics[width=\linewidth]{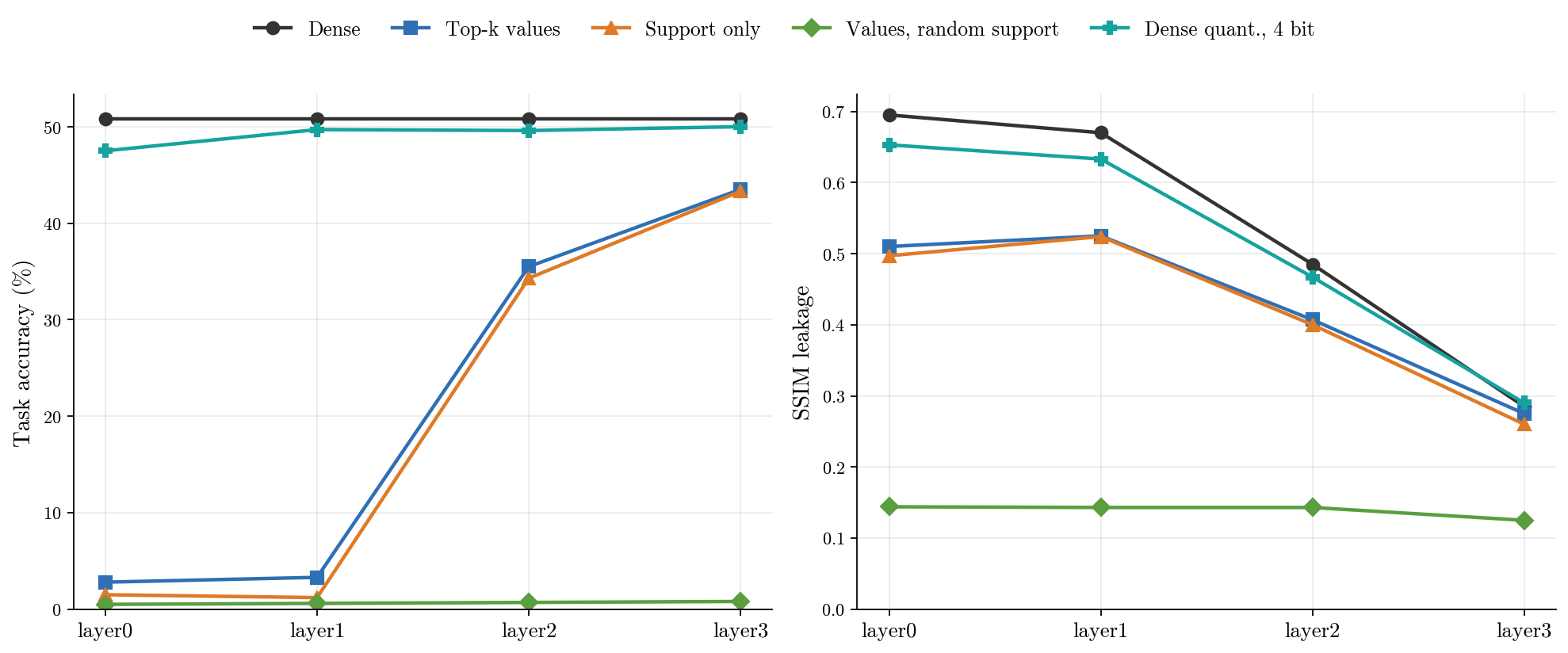}
\caption{Layer ablation (ResNet-18). Task accuracy (left) and SSIM leakage
(right) across split layers 0 to 3. Positions-only follows the top-$k$ activation at
every layer, and values on random positions remain at the floor.}
\label{fig:abl-layer}
\end{subfigure}
\hfill
\begin{subfigure}{0.49\textwidth}
\centering
\includegraphics[width=\linewidth]{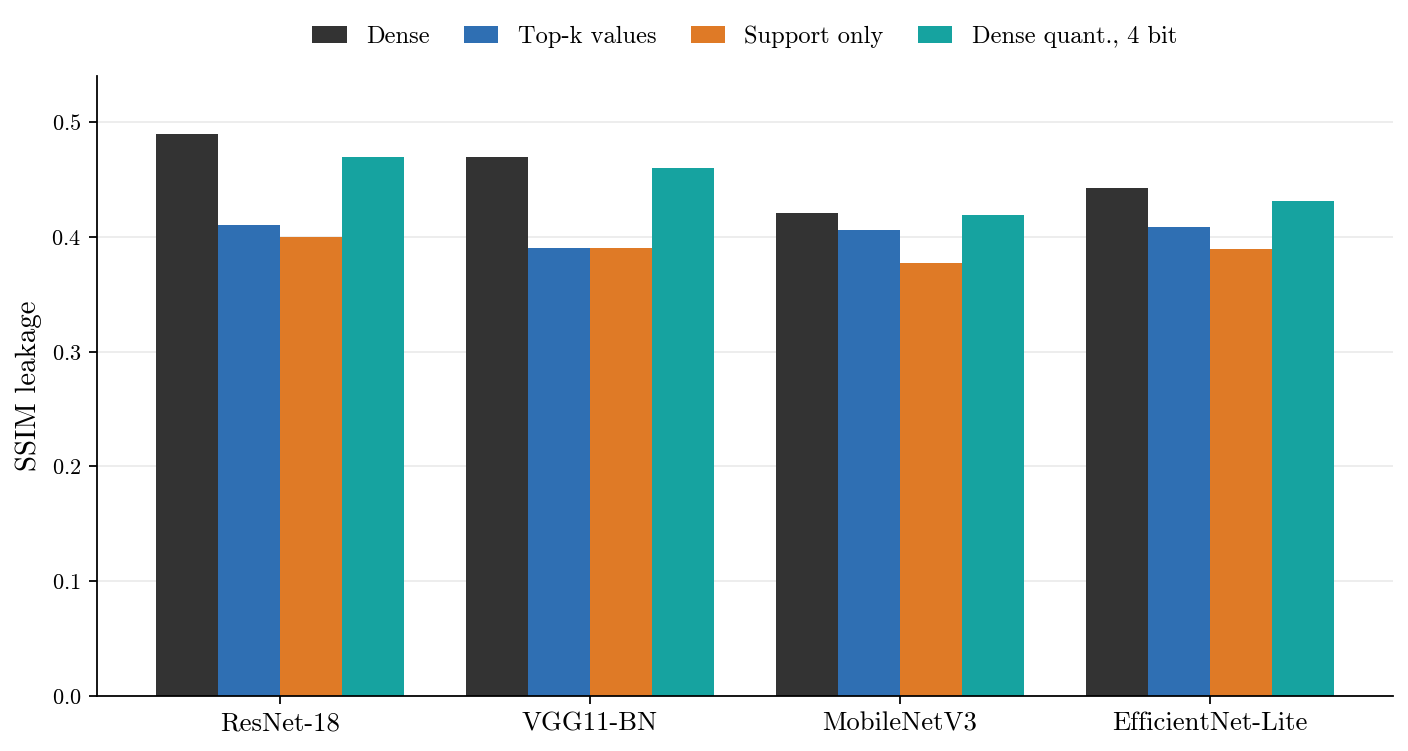}
\caption{Architecture ablation at \texttt{layer2}. Reconstruction SSIM
leakage across four backbones. Positions-only is just below the top-$k$ activation
and follows dense $4$-bit in every backbone.}
\label{fig:abl-arch}
\end{subfigure}
\caption{Robustness of the values-versus-positions result (TinyImageNet, Aux
inverse): (a) across split layers and (b) across backbones.}
\label{fig:ablations}
\end{figure*}

Finally, we test whether the values-versus-positions result is specific to the
split layer or the backbone. Across ResNet-18 layers 0 to 3
(\cref{fig:abl-layer}), positions-only closely follows the full top-$k$
activation in both task accuracy and SSIM leakage at every layer, and values on
random positions remain at the floor throughout, confirming that the recovered
structure rides on the input-dependent positions rather than on the retained
values. Leakage is present at all depths, but usable utility appears only at
deeper splits, so the practical risk concentrates where high leakage and usable
utility coincide. The same ordering holds across ResNet-18, VGG11-BN,
MobileNetV3, and EfficientNet-Lite at \texttt{layer2} (\cref{fig:abl-arch});
absolute leakage levels differ between backbones, so values and positions are
compared within each backbone, and in every backbone positions-only sits just
below the full top-$k$ activation.

\subsubsection{Structured sparsity.}
Throughout we have been discussing isolating positions from values in sparse activations. We established that positions are the source of privacy leakage. However, global top-$k$ has a doubly informative support: it reveals both \emph{which}
channels are active and \emph{where} within each. We therefore repeat the
values-versus-positions contrast under \emph{per-channel} top-$k$, which keeps a
fixed number of units in every channel and so reveals nothing about channel
selection, only the spatial layout inside each channel. Even with this less
informative support the attribution holds: at $95\%$ sparsity, positions-only
reaches $0.368$ SSIM, $94\%$ of the $0.392$ from the full per-channel activation
(the global case gives $97\%$), while values on random positions stay at the
floor ($0.122$). Stripping the channel-selection axis thus leaves the ordering
intact, i.e., the positions still carry the leakage.

\section{Discussion}
\label{sec:discussion}
\subsubsection{The positions are privacy sensitive.}
Which coordinates a top-$k$ activation keeps is fixed by the input, so the set of
kept positions is itself a compact code for the input. Across datasets this code
lets an attacker reconstruct almost as well as the full sparse activation, and on
faces it preserves identity. A collaborative-inference system that protects only
the transmitted values, while still exposing the positions, therefore keeps a
large visual and biometric attack surface. The index stream is not harmless side
information for decoding, and it belongs inside the privacy threat model.

\subsubsection{Identification persists at low rate and low utility.}
We showed that re-identification from positions only is feasible under learned attackers. A positions-only mask
at $95\%$ sparsity sends under a third of a bit per dimension and loses fine
facial detail, yet it still re-identifies unseen subjects at $44\times$ chance,
and the leakage falls only slowly as rate and utility drop further. Low measured
utility is not the same as low information content. The frozen server simply
stops using information that the learned inverse still recovers. Because
reconstructions that re-identify a person can fall within data-protection
regulation such as the GDPR, a compliance assessment should rest on measured
re-identification risk rather than on transmitted volume or pixel fidelity
alone.

\subsubsection{Privacy needs multiple metrics and attackers.}
A single metric or a single attacker can make a representation look safe when it
is not. Pixel similarity is one such metric. For example, SSIM rates the positions-only
reconstructions as poor, yet FaceNet retrieval recovers the subject's identity
from those same reconstructions, so SSIM alone understates the risk. The choice
of attacker matters in the same way. On the discrete mask the white-box audit
recovers little, while the learned inverse recovers faces, so the audit alone
also understates the risk. A privacy claim should therefore be made using both
visual and biometric metrics and under the strongest available attack, which is
the learned inverse whenever the server could plausibly hold auxiliary data.

\subsubsection{Implications for defenses.}
The same probes also point to where a defense has to act. If the positions are
randomized fresh for every input, they stop leaking, but the server can no
longer read them and its accuracy collapses (\cref{tab:combined95}). If instead
the positions are fixed and input-independent, as in a static mask, the server
keeps working, but the real values sent on that fixed support still leak
(\cref{fig:pareto}). Randomized
sparsification~\cite{zheng2023randomized,hoefler2026leveraging} sits between
these two cases and trades one effect off against the other. The conclusion is that
a defense for sparse CI has to be judged on how much its positions leak and not
only on how well it hides its values, and the component-isolated audit we
introduce is the tool for measuring exactly that.

\section{Scope and Limitations}
\label{sec:limitations}
Our experiments use ResNet-18 split at \texttt{layer2}, and \cref{sec:res-ablations}
shows the result is robust across split layers and backbones. Several settings
remain open. These include other structured schemes such as channel pruning and
block and $N{:}M$ sparsity, transformer token pruning, which is itself a
positions-only transmission, and deployed entropy codecs. Our rates are also
analytical upper bounds from \cref{eq:rate} rather than on-the-wire measurements.
Because correlated positions compress further than the bound, this can only
understate the positions' leakage per transmitted bit, and measured codec and
end-to-end wire rates are left to future work. An ethics and data-governance
statement is given in \suppEthics.

\section{Conclusion}
\label{sec:conclusion}
We examined the assumption that sparsifying intermediate activations improves
the privacy of collaborative inference, and found it holds only in limited scenarios. By
transmitting the retained values and their positions in isolation, we showed
that sparsification reduces leakage far less than it reduces the rate, and that
the remaining leakage is carried almost entirely by the positions, which act as
a low-rate representation of the input. On face data the positions alone
preserve most of the reconstruction quality and re-identification performance of
the full sparse activation, at operating points where both the rate and the task
utility are low. Because the standard white-box audit underestimates this
leakage while a learned inverse trained on modest auxiliary data recovers
identity-bearing reconstructions, sparse collaborative inference should be
audited under learned inversion, and the input-dependent positions treated as
sensitive transmitted data throughout the pipeline.
\newpage
\section*{Acknowledgements}
This work was supported by the European Union’s Horizon Europe research and innovation programme (EU Horizon Europe) as grant ACHILLES (101189689).

\bibliographystyle{splncs04}
\bibliography{main}

\clearpage
\appendix
\setcounter{figure}{0}
\setcounter{table}{0}
\renewcommand{\thefigure}{\Alph{section}\arabic{figure}}
\renewcommand{\thetable}{\Alph{section}\arabic{table}}

\noindent This appendix provides the full implementation and evaluation details
referenced from the main text. Complete per-sparsity tables and additional
qualitative reconstructions are provided with the released code at
\url{https://github.com/an7123/Privacy-Study-Sparse-CI}.

\medskip
\noindent\textbf{Numbering.} Appendix sections are lettered
(\eg, \cref{sec:app-threat}), and figures and tables in the appendix are
prefixed with the section letter (Table~A1, and so on). Plain numbered
references (\cref{sec:res-rate}, \cref{tab:combined95}, \cref{fig:pareto})
point to the main text.

\section{Implementation and Evaluation Details}
\label{sec:appendix}

\subsection{Split Model and Threat Model}
\label{sec:app-threat}
We split a standard torchvision ResNet-18 at the output of \texttt{layer2}, so the
client computes $f_{\mathrm{c}} = \{\text{stem}, \texttt{layer1}, \texttt{layer2}\}$
and the server computes $f_{\mathrm{s}} = \{\texttt{layer3}, \texttt{layer4},
\text{avgpool}, \text{fc}\}$. The transmitted activation is taken after the
\texttt{layer2} ReLU and is therefore non-negative, which is what allows the values
and their magnitudes to coincide.

The activation dimension at the split is $d = C \times H \times W$ with $C = 128$ in
all cases, giving $d = 8192$ for TinyImageNet ($128\times8\times8$), $d = 4608$ for
FaceScrub at $48^2$ ($128\times6\times6$), and $d = 51200$ for Imagenette-160
($128\times20\times20$). Top-$k$ selection is global over the entire $C\times H\times
W$ tensor rather than per channel or per location, with ties broken by the
deterministic order of \texttt{torch.topk}. The keep fraction is $\rho = k/d$, so the
headline operating point $\rho = 0.05$ corresponds to $k = 409$ retained coordinates
on TinyImageNet.

Both attackers are white-box with respect to the client. The honest-but-curious
server knows the architecture, the split point, the sparsification rule, and the
transmitted object $o = \mathcal{O}(z)$. The learned inverse additionally queries
$f_{\mathrm{c}}$ to generate training pairs $(\mathcal{O}(f_{\mathrm{c}}(x_j)), x_j)$
from auxiliary data, and the white-box optimizer backpropagates through
$f_{\mathrm{c}}$. At test time the decoder observes only the transmitted object $o$,
never $x$ or $z$.

\subsection{Probes and Server Input}
\label{sec:app-probes}
Each probe maps the activation $z$ to a transmitted object. For the utility
measurement this object is fed to the frozen server $f_{\mathrm{s}}$, where
$m \in \{0,1\}^d$ denotes the top-$k$ mask and $\odot$ is elementwise product.
\cref{tab:app-probes} gives the exact transmitted object and analytical rate for
each probe, organized by the two-factor design of \cref{tab:repr-map}.

\begin{table}[h]
\centering
\caption{Transmitted object and analytical rate for each probe. The binary mask is
fed to the server as literal $0/1$ floats occupying the activation tensor, without
magnitude rescaling.}
\label{tab:app-probes}
\small
\setlength{\tabcolsep}{5pt}
\renewcommand{\arraystretch}{1.2}
\begin{tabular}{@{}lll@{}}
\toprule
Probe & Transmitted object & Rate (bits/dim) \\
\midrule
Dense                       & $z$ & $b = 16$ \\
Top-$k$ activation          & $z \odot m$ & $H_2(\rho) + \rho b$ \\
Positions only              & $m \in \{0,1\}^d$ & $H_2(\rho)$ \\
Values on random pos.\      & values scattered to fresh random positions & $H_2(\rho) + \rho b$ \\
Dense $b$-bit quant.\       & symmetric $b$-bit quantization of $z$ & $b$ \\
Static mask~\cite{hoefler2025relevance} & $z \odot m_{\text{fixed}}$ & $\rho b$ \\
\bottomrule
\end{tabular}
\end{table}

The binary mask is passed to the server without any magnitude rescaling, as literal
$0/1$ floats. This models the deployment in which the client transmits the
index set and the server runs unchanged, and it is the reason positions-only
transmission loses accuracy relative to the top-$k$ activation ($2.4$, $3.7$,
and $11.0$ points on TinyImageNet, Imagenette, and FaceScrub, respectively)
rather than none. Rescaling the mask to the per-channel mean activation
magnitude is a straightforward variant that affects only the utility axis; the
leakage measurements do not involve the server.

\subsection{Server Classifier}
\label{sec:app-server}
The server is the original frozen $f_{\mathrm{s}}$. It is not retrained, fine-tuned,
calibrated, or re-normalized for any representation; the transmitted object is fed to
it directly, and downstream top-1 accuracy is measured on this frozen head. The frozen server is a deliberate
design choice, not a limitation: it measures the utility that an unmodified
deployment actually obtains from each transmitted object, and a central claim
of this work is that the privacy leakage persists even where this measured
utility is low (\cref{sec:res-rate,sec:res-biometric}). Low frozen-server
accuracy signals that the unmodified server no longer exploits the transmitted
information, not that the information is absent; the inversions recover
precisely this latent content, and a server retrained per representation, a
different deployment, could likewise recover part of it. The leakage
measurements themselves do not involve the server.

\subsection{Learned Inverse}
\label{sec:app-inverse}
The learned inverse $g_\phi$ adapts an input head to the transmitted object and then
applies an identical generative trunk that maps a $[256, 8, 8]$ latent to the output
image. For a spatial input $[B, C, H, W]$ the head is $\texttt{Conv2d}(C, 256, 3) \to
\text{BN} \to \text{ReLU}$; for a vector input such as per-channel counts $[B, C]$ it
is a linear map to $256\!\cdot\!8\!\cdot\!8$ followed by a reshape. The shared trunk
is three blocks of $[\text{Upsample}\times 2 \to \texttt{Conv}(256, 256, 3) \to
\text{BN} \to \text{ReLU}]$ followed by $\texttt{Conv}(256, 3, 3) \to \text{Sigmoid}$,
producing an output in $[0,1]$. The output size is eight times the split spatial size,
which is exact for all three datasets ($8\!\to\!64$, $6\!\to\!48$, $20\!\to\!160$).

\begin{table}[h]
\centering
\caption{Parameter counts, so that attacker capacity is explicit and comparable across
input layouts. The generative trunk is identical in every inverse; only the input head
changes.}
\label{tab:app-params}
\small
\setlength{\tabcolsep}{6pt}
\renewcommand{\arraystretch}{1.2}
\begin{tabular}{@{}lr@{}}
\toprule
Inverse & Params \\
\midrule
Shared generative trunk & 1.78\,M \\
ConvDecoder, spatial input ($C{=}128$) & 2.07\,M \\
ConvDecoder, occupancy map ($C{=}1$) & 1.78\,M \\
ConvDecoder, per-channel counts (vector, $C{=}128$) & 3.89\,M \\
Values MLP-on-sorted (\cref{sec:app-payload}) & 40.4\,M \\
Values DeepSets (\cref{sec:app-payload}) & 19.4\,M \\
\bottomrule
\end{tabular}
\end{table}

The inverse is trained with Adam at a learning rate of $2\times10^{-3}$, no
learning-rate schedule, an MSE pixel loss, and batch size $256$. The main TinyImageNet
models train for $12$ epochs (with $6$ for sanity runs and $12$ to $18$ for the
qualitative figures) and the FaceScrub models for $20$ epochs. The architecture,
capacity, and optimization budget are identical across all probes, and only the
channel count of the input head changes. By default the inverse is trained on the full
training split (TinyImageNet $100$k; FaceScrub, the decoder's identity subset). The
auxiliary-set sweep covers sizes $\{100, 500, 2000, 10000, 100000\}$ at a constant
budget of approximately $4000$ gradient steps, with small sets repeating their data, so
that the sweep isolates the amount of data rather than the length of training.

\subsection{White-Box Inversion}
\label{sec:app-wb}
The white-box attacker minimizes
\begin{equation}
\hat{x} \;=\; \arg\min_{x'}\;
\ell\bigl(\mathcal{O}(f_{\mathrm{c}}(x')),\, o\bigr)
\;+\; \lambda_{\text{tv}}\,\mathrm{TV}(x')
\;+\; \lambda_{\ell_2}\,\lVert x' - \tfrac{1}{2}\mathbf{1} \rVert_2^2 .
\label{eq:app-wb}
\end{equation}
Write $a' = f_{\mathrm{c}}(x')$ for the activation of the candidate image and
$\langle v \rangle_{\mathcal{I}} = \tfrac{1}{|\mathcal{I}|}\sum_{i\in\mathcal{I}} v_i$
for the mean of a vector $v$ over an index set $\mathcal{I}$. For continuous payloads
and for dense activations, the fidelity term is the masked mean-squared error on the
transmitted positions,
\begin{equation}
\ell_{\text{cont}}(a', o) \;=\;
\frac{\lVert (a' - o)\odot m \rVert_2^2}{\lVert m \rVert_1} .
\label{eq:app-wb-cont}
\end{equation}
For a binary positions-only observation the values are unavailable, so $\ell$ is
instead a ranking margin that drives the $k$ largest coordinates of $a'$ onto the
observed mask,
\begin{equation}
\ell_{\text{pos}}(a', m) \;=\;
-\,\bigl\langle a' \bigr\rangle_{\mathcal{S}}
\;+\;
\bigl\langle a' \bigr\rangle_{\mathcal{T}_k(a' \odot (1-m))} ,
\label{eq:app-wb-pos}
\end{equation}
where $\mathcal{S} = \{i : m_i = 1\}$ is the observed set of positions and
$\mathcal{T}_k(\cdot)$ returns the indices of the $k$ largest-magnitude entries of its
argument, here the largest off-support activations. Minimizing $\ell_{\text{pos}}$
raises the on-support activations and suppresses the strongest off-support ones,
without penalizing the unobserved values. The image prior combines total variation
with an $\ell_2$ pull toward the gray image $\tfrac{1}{2}\mathbf{1}$. We optimize the
image directly with Adam at learning rate $0.05$ and $\beta = (0.9, 0.999)$ for $1500$
to $2000$ steps, from the initialization $x = 0.1\,\varepsilon + \tfrac{1}{2}\mathbf{1}$
with $\varepsilon \sim \mathcal{N}(0, I)$ clamped to $[0,1]$, and with
$\lambda_{\text{tv}} = 5\times10^{-2}$ and $\lambda_{\ell_2} = 5\times10^{-4}$,
returning the best-loss iterate.

\subsection{Values-Only Attackers}
\label{sec:app-payload}
To rule out an attacker-architecture mismatch, the values-only attackers share the
identical $1.78$\,M generative trunk and differ only in an input encoder suited to an
unordered set of values. The MLP-on-sorted-values encoder is
$\texttt{BatchNorm1d}(k) \to \text{MLP}(k \to 2048 \to 2048 \to 256\!\cdot\!8\!\cdot\!8)
\to \text{trunk}$. Because the split is post-ReLU, the retained values are non-negative
and their sorted vector is a sufficient statistic for the multiset of values, so this
encoder has access to all information available to any permutation-invariant attacker
on the values. The DeepSets encoder applies a per-element map $\phi(v_i)$, a
permutation-invariant pooling (mean, max, and sum), and a projection to the latent.
Both encoders have at least nine times the parameter count of the positions attacker
($40.4$\,M and $19.4$\,M against $2.07$\,M), so a failure to reconstruct cannot be
attributed to insufficient capacity. We report the unconditional mean-image SSIM as the
zero-information floor.

\subsection{Datasets: Partitions, Preprocessing, Provenance}
\label{sec:app-data}
\textbf{TinyImageNet-200.} $100$k training images ($500$ per class) and $10$k
validation images at $64\times64$. Preprocessing is $\texttt{Resize}(64) \to
\texttt{CenterCrop}(64) \to \texttt{ToTensor}$ into $[0,1]$, followed by ImageNet
normalization (mean $[.485, .456, .406]$, std $[.229, .224, .225]$) before
$f_{\mathrm{c}}$. Reconstruction metrics are computed in the $[0,1]$ space. The
evaluation set is a fixed $1000$ validation images (global seed $12345$), identical
across all probes and attackers.

\textbf{FaceScrub.} A public $48$-pixel crop mirror (faces aligned and cropped offline;
$526$ identities, $41{,}425$ training images and a held-out validation split), resized
to $48\times48$, converted to a tensor, and ImageNet-normalized. The split classifier
is trained on all identities (SGD, learning rate $0.1$, cosine schedule, $30$ epochs,
weight decay $5\times10^{-4}$), reaching $76.5\%$ validation top-1 over $526$
identities (chance $0.19\%$). For the main re-identification results (\cref{tab:facescrub}) the inverse is
trained on the FaceScrub training split, and retrieval uses a gallery of one
held-out image for each of the $211$ evaluation identities, with each query
being the reconstruction of a different held-out image of the same identity.
Query and gallery images are therefore always distinct, while identities may
overlap between decoder training and the gallery. The identity-disjoint protocol
of \cref{sec:res-biometric} instead partitions identities into a
decoder-training set ($60\%$) and a disjoint evaluation pool ($40\%$) that
supplies both the queries and the one-image-per-identity gallery, with a fresh
partition per seed, so no evaluated identity is ever seen in training. An earlier processed copy of
FaceScrub became unavailable and the dataset was re-acquired from a public crop
mirror; this lowers the absolute FaceNet cosine (from $0.569$ to $0.508$) but
leaves SSIM, rank-1, and every relative conclusion unchanged.

\textbf{Imagenette-160.} Images at $160\times160$ ($d = 51200$), used as the
higher-resolution check.

\subsection{Matched Attacker Capacity Across Input Layouts}
\label{sec:app-capacity}
Capacity is matched at the generative trunk, which has an identical $1.78$\,M
parameters in every inverse. The input head adapts to the transmitted object, a
$1\times1$ or $3\times3$ convolution for spatial masks and a linear map for vectors or
sets, and for the values-only set attackers the input encoder is sized at or above the
positions attacker's total (\cref{sec:app-payload}). No representation is therefore
handicapped by decoder capacity; where the input layouts differ, the attacker suited to
that layout is given more parameters, not fewer.

\subsection{Positional Structure Ablation}
\label{sec:app-structure}
We degrade one axis of the support at a time at fixed density. Destroying
channel identity while preserving each spatial plane (a channel shuffle) drops
the leakage to the random-positions floor ($0.136$ versus $0.428$ SSIM for the
intact support), as does preserving channel identity while permuting locations
within each plane (a spatial shuffle, $0.132$); reducing the support to
per-channel active counts, with all locations discarded, is likewise near the
floor ($0.158$). The leakage therefore requires the \emph{joint}
channel-and-spatial arrangement of the retained units, and in particular the
spatial layout alone does not carry it, so the positional leakage is not a
two-dimensional saliency map.

\subsection{Conditional Contribution of the Values}
\label{sec:app-increment}
Given the correct support, the retained values add only a small increment over
positions-only: $0.007$ SSIM on TinyImageNet and $0.020$ on Imagenette
(\cref{tab:combined95}), and $0.4$ rank-1 points on FaceScrub
(\cref{tab:facescrub}). Placed on random positions the same values fall to the
dataset floor ($0.145$ SSIM), so what they lack is \emph{unconditional}
information about the input rather than all information; the positions are
sufficient for most of the leakage, not the sole possible source of it.

\subsection{The Static Mask in Detail}
\label{sec:app-staticmask}
The static mask of~\cite{hoefler2025relevance} is the input-independent endpoint
of this design space, and it is instructive about what input-independence does
and does not buy. Its fixed support carries no positional information, yet it
still transmits real values on that support and therefore still leaks
(\cref{fig:pareto}): input-independence removes the \emph{position} channel, not
the \emph{value} channel. Only values on a \emph{per-sample random} support
collapse to the floor (\cref{tab:combined95}), because there no fixed
value-to-pixel mapping can be learned. What the static mask lacks is
frozen-server utility: at rate $3.2$ it reaches only $11\%$ accuracy (against
$44\%$ for dynamic positions at rate $0.47$), since a fixed support discards the
input-specific information the unmodified server requires. Adapting the server
is a different deployment that may recover utility; wherever dynamic top-$k$ is
chosen for its compression and utility, its input-dependent positions belong
within the privacy threat model.

\subsection{Ethics and Data Governance}
\label{sec:app-ethics}
This study applies established feature-inversion methods to quantify a
previously under-measured leakage channel. FaceScrub is used only within the
closed re-identification benchmark, and the trained inverses and reconstructions
are retained solely for evaluation, not released as an identification tool. We
consider the measurement protective on balance, since it allows system builders
to treat the index stream as sensitive rather than as harmless side
information.

\end{document}